\documentclass{iopjournal}

\usepackage{dcolumn}
\usepackage{bm}
\usepackage{amssymb,amsmath} %
\usepackage{graphicx}
\usepackage{hyperref}
\usepackage{xcolor}

\begin{document}
\articletype{Paper} 

\title{Coordination Requires Simplification: Thermodynamic Bounds on Multi-Objective Compromise in Natural and Artificial Intelligence}

\author{Atma Anand$^1$\orcid{0000-0003-0616-7955}}

\affil{$^1$Department of Physics and Astronomy, \\ University of Rochester, Rochester, NY USA}

\email{atma.anand@rochester.edu}

\keywords{Scaling laws of complex systems, Thermodynamics of Computation, Information Theory, Statistical Mechanics, Chaos \& nonlinear dynamics, Artificial Intelligence, Social Physics}

\date{\today}

\begin{abstract}
Information-processing systems that coordinate multiple agents and objectives face fundamental thermodynamic constraints. We show that solutions with maximum utility to act as coordination focal points have a much higher selection pressure for being findable across agents rather than accuracy. We derive that the information-theoretic minimum description length of coordination protocols to precision $\varepsilon$ scales as $L(P)\geq NK\log_2 K+N^2d^2\log (1/\varepsilon)$ for $N$ agents with $d$ potentially conflicting objectives and internal model complexity $K$. This scaling forces progressive simplification, with coordination dynamics changing the environment itself and shifting optimization across hierarchical levels. Moving from established focal points requires re-coordination, creating persistent metastable states and hysteresis until significant environmental shifts trigger phase transitions through spontaneous symmetry breaking. We operationally define coordination temperature to predict critical phenomena and estimate coordination work costs, identifying measurable signatures across systems from neural networks to restaurant bills to bureaucracies. Extending the topological version of Arrow's theorem on the impossibility of consistent preference aggregation, we find it recursively binds whenever preferences are combined. This potentially explains the indefinite cycling in multi-objective gradient descent and alignment faking in Large Language Models trained with reinforcement learning with human feedback. We term this framework Thermodynamic Coordination Theory (TCT), which demonstrates that coordination requires radical information loss.
\end{abstract}


\section{Introduction}
Agents in complex systems face fundamental information-theoretic constraints as finite resources must model environments of arbitrary complexity. Working memory limitations force agents to function with a smaller subset of available information, thereby creating a bottleneck~\cite{Cowan2001}. 
While coordination with other agents with different objectives can increase collective capacity, information is both compressed and lost at every emergent hierarchical level~\cite{libby2007noisy}. This principle that any tractable macroscopic description requires information loss by coarse-graining over microscopic degrees of freedom, has deep parallels in the quantum origins of the classical world~\cite{zurek2003decoherence}. We elaborate how this establishes a fundamental tension between scaling and accuracy.
The Free Energy Principle (FEP) describes how individual systems minimize surprise by updating internal models using prediction errors from sensory inputs through Bayesian inference~\cite{friston2010free,friston2013life}, but no current formalisms address the information-theoretic costs and emergent behavior when multiple agents must coordinate their predictive models across potentially conflicting objectives. This is the gap this paper addresses.

Anderson showed `more is different'~\cite{Anderson1972}, i.e., collective properties emerge at scale that cannot be reduced to constituent behaviors. We quantify this emergence mechanism for information-processing systems: coordination requires radical simplification because agents must converge on Schelling focal points~\cite{schelling1960strategy} despite having conflicting objectives and limited communication capacity. This convergence forces systematic information loss. When selection pressures for findability, accuracy, and other objectives compete, emergent dynamics determine which information the system preserves and discards.

This framework builds upon computational thermodynamic results to establish our scaling bounds. Landauer's principle provides the minimum energy required for information erasure~\cite{landauer1961irreversibility}. Bennett extended this thermodynamic cost to general computation~\cite{bennett1982thermodynamics}. Wolpert's synthesis connects stochastic thermodynamics to computation theory~\cite{wolpert2019thermo}, while the no free lunch theorems~\cite{wolpert1997no} and works on physical limits of inference~\cite{wolpert2008physical} establish that no universal optimization exists. We leverage these results to prove multi-objective coordination cannot have a general solution and that coordination has thermodynamic costs that scale super-linearly. 
In parallel, information-theoretic approaches to prediction~\cite{still2012thermodynamics} and decision-making~\cite{tishby2011information} establish optimal behavior and energetic costs that we will show multiply when agents must coordinate. Information theory~\cite{Cover2006} provides the mathematical foundation for our quantification of these costs.

The theoretical foundation also rests on computational impossibility results that constrain achievable coordination. The free energy requirements of biological organisms demonstrate that maintaining organization against entropy imposes unavoidable metabolic costs~\cite{wolpert2016free}. These combine with coordination limitations from Hayek's knowledge problem which identified the incompressibility of distributed information~\cite{hayek1945use}. Ashby's Law of Requisite Variety states that controllers must match environmental complexity~\cite{ashby1956introduction}, but we will show this becomes impossible as $N$ increases. Dunbar's number provides empirical evidence for coordination limits in human groups~\cite{dunbar1992neocortex}, which we explain as the point where our derived communication protocol length vastly exceeds collective working memory. 

Arrow's impossibility theorem~\cite{Arrow1963} proves no perfect preference aggregation exists, which Chichilnisky extended topologically~\cite{Chichilnisky1980} to continuous preference spaces.
Chichilnisky showed that continuity and anonymity axioms can enable possibility results. However, these axioms require homogeneous agents with smooth preferences, a condition violated in most real-world scenarios involving heterogeneous agents coordinating across discrete objectives. We use the impossibility aspect of this topological framework for an alternative derivation of our Theorem 1.

Recent work has approached the multi-agent FEP from various angles. Constant et al. examined regimes of social expectations~\cite{constant2019regimes}, focusing on conformity rather than the accuracy tradeoffs we quantify. Ramstead et al. explored cultural affordances as scaffolding~\cite{ramstead2016cultural}, providing qualitative insights we now formalize. Vasil et al.~\cite{vasil2020world} treated communication as active inference, but here we will show that perfect communication is impossible at scale. Predictive processing frameworks~\cite{clark2013whatever} explain perception as prediction error minimization but have not been extended to multiple agents and conflicting objectives. Further, we identify phase transitions to different focal points at critical ``coordination temperatures'' that are suggested by our scaling laws and theorems. These theoretical predictions align with recent observations in machine learning, where multi-objective gradient descent exhibits indefinite cycling~\cite{Yu2020,Liu2021}, suggesting our bounds are fundamental rather than algorithmic.

We explicitly consider classical systems making quantum effects such as entanglement-assisted coordination outside our scope. We analyze open systems coupled to environmental energy reservoirs, deriving how external flux modifies coordination bounds. We term this framework Thermodynamic Coordination Theory (TCT), which establishes fundamental limits on achievable multi-agent, multi-objective coordination accuracy. We show that coordination costs scale super-linearly even with parallel communication, and that findability pressure on utility dominates accuracy, define operational and measurement protocols for theoretical quantities, identify scaling laws and phase transitions near critical points, explain observed Large Language Model (LLM) behavior and other complex systems, and discuss falsification criteria.

This paper synthesizes insights from multiple fields but is largely self-contained. While we cite literature from statistical mechanics, information theory, computer science, and social choice theory for context, the main arguments require only basic familiarity with information theory (measured in bits) and probability. Readers from different backgrounds may find some sections more intuitive than others. The rest of the paper is organized as follows. In Sec. 2, we illustrate our theoretical framework with the example of bill-splitting or ``bistromathics''. The main results and their derivations are provided in Sec. 3. The expected critical phenomena and physical parameters are described in Sec. 4. We conclude and discuss the implications of TCT in Sec. 5.

\section{Restaurant Bill Splitting as Illustration}\label{sec:bistro}

To understand why coordination forces simplification, consider the familiar example of $N$ diners attempting to split a restaurant bill in accordance with $d$ objectives. This simple problem exhibits the same information-theoretic constraints that govern all multi-agent coordination.

\textit{Sample Bill-Splitting Problem}: $N$ agents must coordinate payment for shared consumption with $d$ constraints. For example, agents may choose $d=1$: equally split the total amount (protocol communication dominates), $d=2$: individual items and fairness, 
$d=3$: add who could not eat what (limited quantities, dietary restrictions, etc.), $d=4$: add wealth disparities (who can afford what).

Human working memory capacity is limited to approximately $4\pm1$ chunks~\cite{Cowan2001,Cowan2015}, which we estimate for this paper to be $\mathcal{O}(10^2)$ bits when accounting for the information content of meaningful stimuli~\cite{Brady2016}.

For simplicity, we assume each agent's internal model complexity is 10 bits for each objective, and each pairwise resolution to desired cost splitting precision needs 5 bits. 
These numbers are illustrative order of magnitude estimates, and we shall see they quickly exceed human memory capacity ($\sim 100$ bits) for any reasonable values.


For $N=4$ diners to \textit{completely} coordinate with $d=2$ (items and fairness): 
Each diner must communicate items consumed, e.g. ``I had the soup (\$8), half the calamari (\$6), two beers (\$12)'' yields $K \approx 20$ bits (prices, quantities, shares, customizations), 
and $\binom{4}{2} \times \binom{2}{2} = 6$ pairwise reconciliation ($\sim$ 5 bits each), e.g. ``Did we split the calamari evenly?''. Total communication: 80 + 30 = 110 bits. For an explanation of multiplication in the pairwise reconciliation, see the topological obstruction created by preference cycling of objectives described in the topological derivation of Theorem 1 in Supplementary Section 1. 

Note: We use the combinatorial notation $\binom{N}{2}=\frac{N(N-1)}{2}$. In this section, we use the more intuitive $\binom{d}{2}$ for simplicity when discussing pairwise objective conflicts, though the full coordination cost in Theorem 1 includes both the mean vector and covariance matrix elements yielding $\frac{d(d+3)}{2}$.

For $N=8$ with $d=4$ (full constraints): individual consumption data, $K = 8 \times 40 = 320$ bits; pairwise negotiations for all objectives and agents = $\binom{8}{2} \times \binom{4}{2} \times 5 = 840 $ bits; total conflict resolution protocol  $\approx 50$ bits. In total, $320 + 840 + 50 = 1,210$ bits of information is exchanged.

This exceeds human working memory by an order of magnitude, creating a bottleneck that forces dimensional reduction to findable focal points~\cite{schelling1960strategy}, e.g., ``Split evenly'' (ignoring individual consumption), ``Separate checks'' (avoiding coordination entirely), ``I'll get it'' (single payer absorbs complexity costs). When coordination costs exceed perceived utility gains, agents may accept arbitrary errors to avoid coordination. Further, agents become averse to attempting accurate bill-splitting in future dinners preferring their previously coordinated focal points.

This problem termed ``bistromathics'' by Douglas Adams in book two of his five-part \emph{Hitchhiker's Guide to the Galaxy} trilogy shows bill calculations become non-absolute (focal) points~\cite{Adams1980Restaurant}, representing phase transition to simplified coordination. Conflict resolution protocols include restaurants imposing terms like ``3 cards maximum'' or ``20 \% minimum gratuity for parties of 6 or more'', and agents declaring they do not use digital peer-to-peer payments or carry cash. Varying $N$ is qualitatively explored in Section 3 of the Supplementary. Finding the non-absolute focal points (Adams' \textit{recipriversexclusions}) involves Level II chaos~\cite{Harari2015sapiens} which is beyond this paper's scope. 

\textbf{Level II Chaos} (Definition): A term for a specific subclass of Complex Adaptive Systems (CAS), the formal framework for systems of interacting, adaptive agents~\cite{holland1992complex}. Level II chaos describes systems whose trajectories are unknowable because of the necessary recall and loss of information required to transition between states. The subset of information recalled is path-dependent and unknowable a priori, as agents must select which information to preserve when complete description length exceeds system capacity. In these dynamics, agents continuously attempt to update their internal models to predict the system's future state, which in turn alters their and the system's behavior. This reflexive feedback loop is particularly evident in financial markets, where the collective forecasting efforts of participants shape the very price movements they attempt to predict~\cite{arthur1994inductive}. Unlike Level I chaos (e.g., weather, N-body dynamics), whose dynamics are unaffected by prediction, Level II chaos is an emergent property of this process: the unpredictability of macro-state is compounded by the act of trying to forecast it as it changes accessible information. While Level II systems exhibit their own reflexive dynamics, they remain susceptible to Level I influences, for instance, stock returns correlating with local sunshine~\cite{hirshleifer2003good}.

This definition is crucial for our framework as coordination attempts fundamentally alter the landscape being coordinated. Established focal points exhibit metastability and path-dependent hysteresis, making re-coordination thermodynamically costly. While this characterization provides a useful conceptual framework for understanding reflexive dynamics in complex systems, mathematical formalization and empirical validation remain topics for future work. Level II Chaos with the Bistromathics example is illustrated in Section 3 of the Supplementary.

\section{Mathematical Derivations}
This section derives the quadratic $N^2d^2$ scaling term in the complete coordination protocol length (Theorem 1), and findability dominating accuracy whenever coordination increases survival (Theorem 2).

For a given time-bound event of arbitrary duration, consider $N$ classical agents each with finite memory B bits, operating in an environment with Kolmogorov complexity bits. $\mathcal{K}_{\rm env} \gg B$ trying to coordinate around $d$ uniquely defined and potentially conflicting objectives. The system is open and exchanges energy and information with environment at rate $F_{\rm env}$. 
These results stated here are independent of the system Hamiltonian. 
For this event, each agent $i$ uses internal model $m_i$ with practical complexity $K_i \leq B$, where we define $K_i$ operationally as the minimum bits needed to compress agent i's model to reconstruction error $\varepsilon$ using standard algorithms (e.g., Lempel-Ziv)~\cite{Cover2006}.  For consistency, we use the same $\varepsilon$ as the consensus precision throughout the protocol.

\subsection{Theorem 1: Coordination Info Scale as $N^2d^2$}
In the first interaction, agents cannot assume correlation without coordination to discover it. Even if agent models have almost complete overlap, they must exchange sufficient information to identify that overlap, which itself requires the full information exchange. Only after this initial coordination can hierarchical structures or factions emerge that exploit discovered correlations but shift coordination up a level (see end of section).

\textit{Variable Definitions}: 
$N$ is number of agents which could mean human annotators for AIs or employees for organizations, $K$ is Model complexity in bits such as compressed parameter count for AI or utilized working memory for humans, $\mathcal{K}_{\rm{env}}$ is the environmental Kolmogorov complexity (typically $\gg K$), $d$ is dimensionality of objective space (number of potentially conflicting goals + 1 for ALL approximately compatible goals), $F_{\rm{env}}$ is environmental free energy flux (resources available per unit time), $\rho$ is averaged pairwise model overlap (correlation between agent models, between $0-1$).

\textbf{Definition (Coordination Protocol)}: A coordination protocol $P$ is a set of rules specifying, (i) message generation, $\mu_i: m_i \rightarrow M_i$ mapping agent states to messages, (ii) message processing, $\pi_i: (m_i, \{M_j\}_{j \neq i}) \rightarrow m_i'$ updating states based on received messages, (iii) scheduling which agents communicate when.

The protocol complexity $L(P)$ is the minimum number of bits required to specify $P$ to reconstruction error $\varepsilon$. This equals the protocol's Kolmogorov complexity $\rm{K(P)}$ up to additive constants~\cite{Li2008}. $L(P)$ consists of $L_{\rm{models}} =$ bits needed to specify agent models relevant to coordination, $L_{\rm{comm}}=$ bits that must be communicated between agents, and $L_{\rm{rules}}=$ bits to specify protocol rules. $L_{\rm{rules}} = \mathcal{O}(\log N)$ for large N and neglected for stricter lower bounds, though this term may dominate the simple case of extreme model overlap, $\rho \approx 1$, and $d=1$. $\log$ uses base 2 in this paper unless explicitly stated.

\textbf{Theorem 1 (Multi-dimensional Coordination Protocol Scaling)}: For $N$ heterogeneous agents coordinating across $d$-dimensional objective functions, where each agent $i$ has a model $m_i$ of complexity $\rm{K}_i$ bits, the minimum description length $L(P)$ of any coordination protocol $P$ that achieves $\varepsilon$-approximate consensus satisfies:
\begin{eqnarray} \label{eq:lp}
    L(P) & \geq L_{\rm{models}} + L_{\rm{comm}} \nonumber \\ 
         & = N \bar{K} \log \bar{K} \, h(\rho) + \binom{N}{2} \frac{d(d+3)}{2} \log(1/\varepsilon)
\end{eqnarray}
where, $\bar{K} = \frac{1}{N}\sum_i K_i$ is mean model complexity, $h(\rho) = 1-\rho$
is the approximate model non-overlap factor, and $\varepsilon$ is the coordination precision.

\textbf{Corollary 1:} Even with optimal compression and $d = 2$ objectives, the quadratic term dominates for any practical system size, making coordination information scale as $\mathcal{O}(N^2 d^2)$.

\textbf{Corollary 2:} The model complexity $K(P)$ of coordinated systems is bounded by: min\{environmental complexity, combined protocol expressiveness or $L(P)$, collective working memory ($\leq \rm{BN} $)\}

\textbf{Proof}: We derive each term by analyzing the information-theoretic requirements for coordination.

\textit{Part 1: Model Specification Complexity ($L_{\rm{models}}$)}

For agents to coordinate, they must share sufficient model information to predict each other's objectives. Each agent i has internal model $m_i$ with Kolmogorov complexity $K_i$. When models have overlap $\rho$, the conditional complexity $K(m_i|m_j) = K_i(1-\rho)$. To specify all models given pairwise communication: $L_{\rm{models}} \geq N \bar{K} \, \log \bar{K} \, h(\rho)$. The $\log \bar{K}$ factor arises from addressing or indexing overhead within each agent's model during protocol execution~\cite{Li2008}. 

In practice, model complexity $K$ may depend on both $N$ and $d$. Each additional objective typically requires additional state representation, suggesting K scales at least linearly with $d$~\cite{Roijers2013}, i.e. $K \geq K_b + K_\delta d$ for some base complexity $K_b$ and $K_\delta$ per objective. The Bistromathics example in Section \ref{sec:bistro} uses $K_b=0, \ K_\delta=10$. Additionally, agents may need to model other agents' states, implying $K$ could scale with $N$ for small groups. However, cognitive limits in humans~\cite{Cowan2001}, and bounded rationality forces categorical simplified representations rather than individual models at large $N$~\cite{Simon1955}. Here we treat $K$ as largely constant to derive lower bounds, noting that any dependence $K(N,d)$ would only worsen the coordination challenge. 

Even with constant model complexity, additional complexity emerges when $K$ exceeds working memory capacity $B$, as agents face an internal coordination problem analogous to multi-agent coordination. The system must aggregate preferences from model components that cannot be simultaneously evaluated, potentially encountering Arrow-like impossibilities in selecting optimal subsets of the internal model. This suggests the $\log K$ addressing overhead represents not just indexing costs but the cost of preference aggregation across model fragments.

\textit{Part 2: Communication Complexity ($L_{\rm{comm}}$)}

For $d$-dimensional objective communication, agents must resolve: 

\textit{Local objective representation}: Each agent i estimates local objective function $R_i: \mathbb{R}^d \rightarrow \mathbb{R}$ from finite samples. Minimal representation assuming a Gaussian model (or any 2 parameter model in general) to precision $ \log(1/\varepsilon) $ bits requires:

Mean vector $\mu_i \in \mathbb{R}^d$: $d \ \log(1/\varepsilon)$ bits

Covariance matrix $\Sigma_i \in \mathbb{R}^{d \times d}$, with diagonal and symmetric off-diagonal elements needing:
\begin{eqnarray}
    d + \binom{d}{2} = \frac{d(d+1)}{2} \log(1/\varepsilon) \text{bits}
\end{eqnarray}

Total per agent is thus $\mathcal{O}\{d^2 \log(1/\varepsilon)\}$ bits.

\textit{Basis alignment}: Even with identical objectives, agents must agree on (i) Coordinate system = $d!$ permutations $\approx d \log d $ bits, (ii) Sign conventions: $2^d$ choices $= d $ bits, (iii) Degenerate subspace handling: $\mathcal{O}(d^2)$ bits worst case (excluded from calculation).

\textit{Pairwise communication}: Using information-theoretic minimum optimal protocols with no fault tolerance overheads, the bits each pair of agents must exchange come primarily from the mean and covariance of models:
\begin{equation}
    I_{pair} = \underbrace{\frac{d(d+3)}{2} \log(1/\varepsilon)}_{\text{objective params}} + \underbrace{d \log d}_{\text{basis alignment}} + \underbrace{\mathcal{O}(d)}_{\text{overhead}}
\end{equation}

For pairwise resolution with N agents:
\begin{eqnarray}
    L_{\rm{comm}} & \geq \frac{N(N-1)}{2} \cdot I_{pair} \nonumber \\
    & = \binom{N}{2} \frac{d(d+3)}{2} \log(1/\varepsilon) \cdot \{1 + \mathcal{O}(1/d)\} 
\end{eqnarray}


\textit{Combining terms}:
$$L(P) \geq L_{\rm{models}} + L_{\rm{comm}} + L_{\rm{rules}}$$
\begin{equation}
    L(P) \geq N \bar{K} \log \bar{K} \, h(\rho) + \binom{N}{2} \frac{d(d+3)}{2} \log(1/\varepsilon)
\end{equation}


For large $N$, if $\bar{K}$ saturates ($ \approx B=100$) and $\rho$ is high $(=0.9 \text{ e.g.)}$, the second term dominates even when $d = 2$. With practical values, $N = 100$, $d = 2$, $\varepsilon = 0.01$:

\begin{eqnarray}\label{eq:t1comp}
   L_{\mathrm{models}} \geq & 100 \cdot 100 \cdot \log(100) \cdot 0.1 \approx 6.6 \times 10^3 \text{ bits} \nonumber \\
   L_{\mathrm{comm}} \geq & \frac{100 \cdot 99}{2} \cdot 5 \cdot \log(200) \approx 7.6 \times 10^4 \text{ bits}
\end{eqnarray}


We provide an independent derivation of Theorem 1 via the topology of preference spaces from the Chichilnisky extension of Arrow's Impossibility theorem~\cite{Chichilnisky1980}, confirming the $\rm{N^2d^2}$ scaling through geometric reasoning (Supplementary Section 1). This convergence of information-theoretic and topological approaches suggests the scaling is fundamental, not algorithmic. Therefore, no protocol innovation can reduce this bound without sacrificing coordination accuracy. Figure \ref{fig:tct_main} panel (a) shows a graphical representation of how $L(P)$ scales with $N$ and $d$.

\textbf{Why Scaling as $N^2d^2$ is Unavoidable:} Alternative aggregation approaches might avoid quadratic scaling but be unable to maintain the accuracy needed for complete coordination.

\textit{Star topology} (centralized): Reduces messages to $\mathcal{O}(N)$ but (i) Each center must process $N \cdot d^2 \log(1/\varepsilon)$ bits, (ii) Single point of failure requires redundancy requiring $\mathcal{O}(N^2)$ bits, (iii) Byzantine fault tolerance requires $f < N/3$ confirmations~\cite{lamport1982byzantine}.

\textit{Sampling/sketching}: Reduces communication but Johnson-Lindenstrauss requires $\mathcal{O}(\log N/\varepsilon^2)$ dimensions~\cite{johnson1984extensions}. As we show using Chichilnisky's theorem, for $d$ objectives with potential conflicts, no consistent sketch exists. Adversarial agents can exploit sampling similar to Byzantine generals problem.

\textit{Hierarchical aggregation}: With branching factor $b$ and depth $\log_b N$ (i) Information loss per level $= \varepsilon_{\text{level}} = 1-(1-\varepsilon)^{1/\log_b N}$, (ii) Total error: $\varepsilon_{\text{total}} = 1-(1-\varepsilon_{\text{level}})^{\log_b N} \approx \varepsilon \log_b N$. To maintain target $\varepsilon$, must reduce per-level error reverting to $\mathcal{O}(N^2)$ bits (see below).

\textit{Information in Hierarchical Coordination:}
To further address whether hierarchical organization can overcome the $N^2d^2$ scaling, consider $N$ agents organized into $M$ groups of size $N/M$ each. Let $d_{\rm{H}} \leq d$ represents the reduced dimensionality of objectives at hierarchical boundaries and within them. Assuming $K$ also remains the same within and between groups, the total coordination length is:
\begin{eqnarray}
L_{\text{H}} & = (N+M)\bar{K}\log\bar{K} \, h(\rho) + \\
& \left[ \frac{N^2 - NM}{2M} + \frac{M(M-1)}{2} \right] \frac{d_{\rm{H}}(d_{\rm{H}}+3)}{2}\log(1/\varepsilon) \nonumber
\end{eqnarray}
where the first term represents model sharing, the second within-group coordination, and the third between-group coordination. See Supplementary Section 2 for full derivations. Minimizing in the limit of large $N$ over $M$ yields $M_{opt} = N^{2/3}$ and:
\begin{equation}
L_{\text{H}}^{min} \approx N\bar{K}\log\bar{K} \, h(\rho) + \frac{1}{2}N^{4/3}d_{\rm{H}}(d_{\rm{H}}+3)\log(1/\varepsilon)
\end{equation}

This reduces communication scaling from $\mathcal{O}(N^2)$ to $\mathcal{O}(N^{4/3})$ but remains superlinear. Since typically $d_{\rm{H}} < d$, $L_{\text{H}}$ can be much lower than $L(P)$, but it comes at the tradeoff of abandoning complete coordination. 

Multi-level hierarchies with branching factor $b$ and depth $\log_b(N)$ can theoretically achieve:
\begin{eqnarray}\label{eq:hier}
L_{\rm{multi}} & \approx  N\bar{K}\log\bar{K} \, h(\rho)  \frac{b^2}{b-1} \nonumber \\
&+ \frac{1}{4}Nb^2d_{\rm{H}}(d_{\rm{H}}+3)\log(1/\varepsilon)
\end{eqnarray}

However, this assumes perfect information aggregation at each level, which Arrow-Chichilnisky's theorem shows is impossible. In practice, hierarchy transforms but does not eliminate the fundamental information scaling constraint.


\subsection{Theorem 2: Findability Dominates Accuracy}
\textbf{Theorem 2:} In multi-agent coordination with utility $U = \Omega[A] \cdot \prod F_i$, the selection pressure for findability exceeds accuracy by factor $\Omega[A]/(F_i \cdot \Omega'[A])$, diverging at accuracy extrema.

\textit{Proof:} Consider a system where at least $M$ stakeholders or emergent factions out of $N$ agents need to accept solution $s \in S$ in the solution space for it to have any utility, where $N > M \gg 1$ for large systems, similar to voting thresholds or quorum requirements.

We define $F_i(s) \in [0,1]$ as the probability of agent $i$ finding and accepting solution $s$, and $A(s) \in [0,1]$ as the accuracy of solution $s$. We are agnostic about the exact normalization across $d$ objectives needed to calculate $A(s)$ in this work.

The probability of system-wide coordination on $s$ is:
\begin{equation}
P(\text{coordinate on } s) = P(s)= \prod_{i=1}^{M} F_i(s)
\end{equation}
which decays almost exponentially as $M$ increases, while $A(s)$ remains constant with respect to $M$.

We define the utility function as:
\begin{equation}
U(s) = \Omega[A(s),M,N] \cdot P(\text{coordinate on } s)
\end{equation}
where $\Omega: \{[0,1],\mathbb{N},\mathbb{N}\} \rightarrow \mathbb{R}^+$ is an unspecified monotonic function of $A$ weighting accuracy and agent thresholds. This multiplicative form follows from network reliability theory~\cite{barlow1975statistical} where system utility requires both solution quality and successful propagation through the agent network, analogous to series reliability or bootstrap percolation~\cite{Watts2002}. Note that this formulation assumes independent discovery and acceptance probabilities, though correlations could be incorporated. While other utility functions like additive or threshold models exist, they might be less suited for describing systems where emergence and collective action are paramount. In such systems, a solution with zero findability or veto power must have zero system-wide utility, a condition that the multiplicative form with stakeholder thresholds naturally captures.

The selection pressure or marginal utility on findability for agent $i$ is:
\begin{equation}
\frac{\partial U}{\partial F_i} = \Omega[A(s)] \cdot \prod_{j \neq i} F_j(s); \ \lim_{F_i \to 1} \frac{\partial U}{\partial F_i} = \Omega \cdot P(s)
\end{equation}
which is independent of $F_i(s)$! As $F_i(s) \rightarrow 1$, agent i enters an ``absorbing state''\footnote{a state where all future time evolution keeps the agent locked in the same region of phase space} by being locked onto \textit{s}. Note that this represents a dynamic process: as agents find and accept \textit{s}, the only way to increase marginal utility of $s$ is to recruit remaining agents, potentially triggering cascade dynamics. This is shown in Figure \ref{fig:tct_main} panel (b) where each cell represents an agent and the color represents which solutions each agent can accept.

The selection pressure on accuracy is:
\begin{equation}
\frac{\partial U}{\partial A} = \Omega'[A(s)] \cdot P(\text{coordinate on } s)
\end{equation}

Therefore, at or near any local or global extrema of $\Omega[A(s)]$ where $\Omega'[A(s)] \rightarrow 0$ the ratio of selection pressures is:
\begin{equation}
\frac{\partial U/\partial F_i}{\partial U/\partial A} = \frac{\Omega[A(s)]}{F_i(s) \cdot \Omega'[A(s)]} \gg 1
\end{equation}

This demonstrates that findability dominates accuracy in coordination dynamics and also implies systems at perceived accuracy maxima experience overwhelming pressure to simplify and propagate rather than improve accuracy further. Note that $P(\text{coordinate on } s) = 0$ if any $F_i(s) = 0$, as observed in systems with veto power (e.g., UN Security Council). 


\section{Critical Phenomena, and Phase Transitions}


We define measurable parameters expected phenomenologically from our two theorems. The definitions are subject to revisions, and re-derivation from a microscopic Hamiltonian in subsequent work. Empirical measurement presents challenges discussed in Section 5.

\subsection{Coordination Temperature}
We define coordination temperature $T_{co}$ as the variance in the content of each agent's models normalized by the mean model complexity:
\begin{equation}
    T_{co} = \frac{1}{N \bar{K}^2}\sum_{i=1}^{N} ||m_i - \bar{m}||^2
\end{equation}
where $\bar{m} = \frac{1}{N}\sum_i m_i$ is the mean model description. In practice, $m_i$ might be difficult to know precisely and the variance will need to be estimated through system-specific proxies.
This connects to physical temperature through Landauer's principle~\cite{landauer1961irreversibility}: each bit of agent's model and coordination protocol requires minimum energy $k_B T_{\text{physical}} \ln(2) $ to erase during coordination. The actual energy costs will be much higher due to inefficiencies and overheads.

A higher $T_{co}$ implies a more disordered system (also see Figure \ref{fig:tct_main}c), in which the findability pressure would lead to a simpler coordination focal point. 
Physical interpretation will be system-specific. For example in neural networks, $T_{co}$ could corresponds to layer activation variance, and bid-ask spread normalized by price for markets.

\subsection{Renormalization Group Flow}
We propose a schematic renormalization group analysis showing system flow toward simplified models:
\begin{equation}
    \frac{dK}{dl} = -y_K(K - K_0)
\end{equation}
where $K$ is model complexity (bits), $l$ is the dimensionless RG flow parameter (number of coarse-graining steps), $K_0$ is the simplest stable coordination focal point (bits), and $y_K > 0$ is the dimensionless scaling exponent. This attractive fixed point suggests repeated coordination drives $K \rightarrow K_0$, representing the minimal model complexity maintaining system viability, or the ``focal point'' selected by thermodynamic pressure.

\subsection{Critical Transitions}
While a rigorous derivation is beyond our current scope, we propose a plausible scaling relation for the critical temperature where systems undergo phase transition to simplified solutions based on the system's key parameters:
\begin{equation}
    T_{c,co} = \frac{K_0/\bar{K}}{\log(N)}
\end{equation}

The $\log(N)$ scaling emerges from hierarchical coordination (Equation \ref{eq:hier}) requiring $\mathcal{O}\{\log(N)\}$ levels, each introducing noise. This is analogous to how network diameter scales as $\log(N)$ in small-world networks. The RG flow becomes unstable when thermal fluctuations exceed the per-agent capacity to maintain coordination, yielding the critical temperature. Systems exhibit metastability, persisting at suboptimal focal points until fluctuations overcome coordination barriers.
The focal points in this framework are analogous to priors in predictive processing and Bayesian brain models~\cite{clark2013whatever, friston2010free}. Both represent informational attractors that minimize surprise through compression.

Beyond $T_{c,co}$, accurate coordination becomes thermodynamically prohibitive. The order parameter $\psi = (K_{absolute} - K_{relative})/K_{absolute}$ measuring model deviation should exhibit rapid transitions near $T_{c,co}$, with width $\Delta T \sim \sqrt{T_{c,co}}$. Here, $K_{absolute}$ is the true system complexity, and $K_{relative}$ is the coordinated model complexity, with $K_{relative} \approx K_0$ at the focal point. This order parameter displays hysteresis characteristic of first-order transitions with groups maintaining $\psi$ values despite crossing $T_{c,co}$ until perturbations trigger reorganization.

Note that the phase transition may occur to a focal point of similar complexity as the immediately preceding one, but it will most likely differ in accuracy. Hybrid phase transitions which combine signature of first order (a sharp, discontinuous change in an order parameter) and second order are also possible.

\subsection{Energy for Metastable Coordination}
From Theorem 2, we expect coordination across larger groups and more objectives requires simpler coordination focal points. However, since each hierarchical level (Equation \ref{eq:hier}) introduces agent model prediction errors due to information loss, complexity cannot spontaneously increase or be maintained at a simpler level without external work.

\begin{figure*}[!htbp]
\centering
\includegraphics[width=\textwidth]{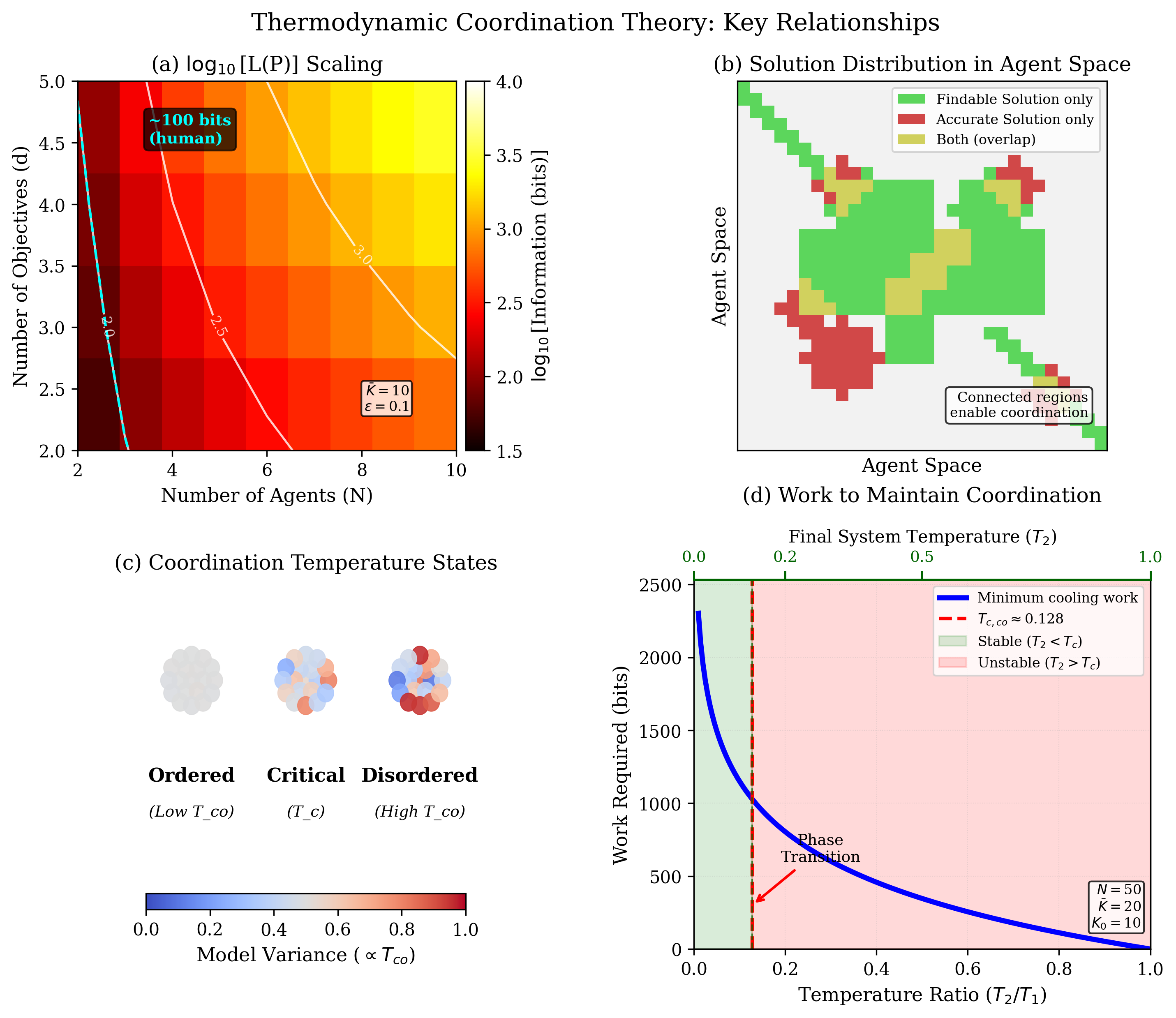}
\caption{Thermodynamic Coordination Theory: Key Relationships. (a) Information scaling showing $\log_{10}$ of coordination protocol length $L(P)$ as a function of number of agents $N$ and objectives $d$. Cyan dashed line marks human working memory limit (100 bits), demonstrating how even modest multi-agent coordination exceeds human capacity. Contours show information growth in orders of magnitude. Parameters: mean agent model complexity $\bar{K}=10$ bits, conflict resolution precision $\varepsilon=0.1$. (b) Distribution of solutions across agent space. Each cell represents one agent and the color represents which solution(s) they can find and agree to implement. Findable solutions (green+yellow) form extended connected regions reaching into corners; accurate but ``invisible'' solutions (red+yellow) exist as isolated islands. Yellow regions indicate where both solution types are available. The green+yellow network forms a spanning cluster enabling system-wide coordination, while red+yellow regions remain disconnected. This collective marginal utility increase drives selection for findable solutions over more accurate one, connecting to topological constraints from Arrow's impossibility theorem (see Supplementary Section 1). (c) Coordination temperature states visualized through variance in agent models. Each circle represents one agent's internal model state. Circular clusters from left to right show: ordered state with tightly aligned models (low $T_{\mathrm{co}}$), critical coordination state with moderate variance, and disordered state with maximal model divergence (high $T_{\mathrm{co}}$). (d) Minimum thermodynamic work required to cool system from initial temperature $T_1$ to final temperature $T_2 (<T_1)$. Bottom axis shows temperature ratio $T_2/T_1$; top axis shows absolute final temperature $T_2$ assuming $T_1=1$. Curve height represents work in bits. Critical temperature $T_{c,\mathrm{co}} \approx 0.128$ determines stability: systems with $T_2 < T_{c,\mathrm{co}}$ (green region) can maintain coordination without additional energy input; systems with $T_2 > T_{c,\mathrm{co}}$ (red region) require continuous resource expenditure for metastable coordination. Parameters: $N=50$, $\bar{K}=20$, $K_0=10$ bits.}
\label{fig:tct_main}
\end{figure*}

Wolpert's no free lunch theorems~\cite{wolpert1997no,Wolpert2023} says no universal optimizer exists across all problems. The TCT description for this would be: no coordination protocol can simultaneously minimize both information loss and energy cost across all agent distributions. To move from coordination temperature $T_1$ to $T_2 (<T_{c,co}< T_1)$ to keep the system stable at the same coordination point, the system has to expend approximate work (in bits):
\begin{equation}
    W \geq N(\bar{K} - K_0) \log(T_2/T_1)
\end{equation}
This is conceptually represented in panel (d) of Figure \ref{fig:tct_main}.

\section{Discussion}
As Thermodynamic Coordination Theory (TCT) appears to span systems across orders of magnitude, we expect and attempt to find signatures across multiple fields and domains.

\subsection{Cross-Domain Signatures}\label{sec:emp}
We identify signatures in existing observations as preliminary validation:

\textit{Artificial Intelligence:} The indefinite cycling observed in multi-objective gradient descent (e.g.~\cite{Yu2020}) directly supports this framework, in particular the extension of Chichilnisky-Arrow's Impossibility Theorem. 
Each loss function $L_i(\theta)$ defines a preference ordering over parameter space, and optimizing $d$ potentially conflicting objectives requires aggregating these orderings into a single gradient update direction, typically through weighted combination: $\nabla\theta = \sum \alpha_i \nabla L_i$. The selection of weights $\{\alpha_i\}$ is topologically equivalent to Chichilnisky's aggregation function $F$: it must map $d$ preference orderings to a collective preference (update direction). The impossibility result applies directly as no continuous, anonymous, unanimous weighting scheme exists when objectives conflict. While different ``gradient surgery'' methods have been proposed to manage conflicting objectives and provide practical improvements, the problem remains algorithmically intractable. The system cannot converge because no consistent aggregation exists, only arbitrary compromises that change under perturbation (e.g.~\cite{noothigattu2018voting}).

Increased computational capacity cannot escape these bounds as larger systems require coordination among more agents (annotators, users, developers) with more objectives, causing $N^2d^2$ to dominate any linear capacity increase, i.e. computational scaling laws do not scale~\cite{Diaz2023}. Reinforcement learning from human feedback could be making the problem worse~\cite{Casper2023}, and leading to alignment faking~\cite{anthropic2024alignment}. Large Language Models (LLM) provide the ideal conditions to test the TCT framework with comparatively simpler ways to measure the parameters needed ($ N, \ d, \ K, \ \varepsilon, \ m_i$). Detailed analysis of these results and their implications for machine learning systems are topics for further work.

\textit{Other Systems:} Qualitative signatures of TCT appear across different contexts and hierarchical levels. 
Human reasoning systematically favors coordination over accuracy~\cite{mercier2017enigma}, with more education paradoxically leading to worse identity protective cognition~\cite{kahan2017misconceptions} to defend focal points. Arrow's impossibility appears to extend to an individual's own preferences, which can be reversed by simple changes in how choices are framed~\cite{tversky1981framing}, suggesting the challenge of consistent preference aggregation arises from the brain's own internal 'society' of competing subsystems even before the multi-agent coordination problem begins~\cite{minsky1986society}.
This observation originally motivated this theoretical framework to explain the similar phenomena in LLMs where scaled systems converge to generic outputs~\cite{anthropic2024alignment,Casper2023,Diaz2023}\footnote{Citations to pre-prints presented at conferences are included due to the rapid pace of research in machine learning.}.
In these contexts, the focal points that emerge from coordination can be understood as the high-level priors that guide the predictive processing of the collective system. 

Cultural evolution selects for transmissible simplifications~\cite{henrich2015secret,boyd1985culture}. Scientific paradigm appears to shift in episodic ``Kuhn cycles'' instead of development by accumulation~\cite{kuhn1962structure}. We predict open source projects are expected to show incremental API complexity decreasing as a power-law: $K_{\rm{API}} \propto N^{-\alpha}$ where $ \alpha \in (0,1)$, reflecting hierarchical coordination requirements (Equation \ref{eq:hier}).

Market bubbles show critical transitions~\cite{mackay1841extraordinary} potentially predictable from trader density, an analysis subject to future work. The difference in the description length of model and communication complexity terms in Equation \ref{eq:t1comp} potentially explains the paradox noted in behavioral finance: why do similarly trained professionals make opposite trades on identical information~\cite{kahneman2021noise}? High model overlap does not imply preference alignment.

\textit{Falsification:} To disprove the predicted scaling relation in Theorem 1 identifying a system working around the coordination costs should be sufficient, i.e., show a large multi-objective system which maintains stability without utilizing resources to keep the coordination temperature low for a timescale comparable to environmental fluctuations or agent ``dropout'' rates. 
For example, Wikipedia might appear to violate the $N^2d^2$ scaling with only $\sim$40,000 active editors coordinating to maintain $\sim$7 million articles, but it is consistent with TCT predictions through non-equilibrium dynamics: editors spend significant resources, averaging 5.8 hours weekly with 12 minutes per edit in 2015~\cite{Bayer2015}, while maintenance backlogs persist with 48,000 unreferenced articles~\cite{WikiProject2025} despite continuous effort~\cite{Halfaker2012}. The system maintains complexity only through constant work against thermodynamic pressure to simplify.

Other falsifiability criteria is a topic for further work beyond the present scope. These patterns suggest thermodynamic necessity: systems with bounded resources must sacrifice accuracy for coordinability when interfacing with unbounded environments. Domain specific empirical analysis will be presented in future work.

\subsection{Scale Free Implications}
The framework suggests that multiple observations across scales are system specific manifestations of the same phenomena. In AI systems, TCT potentially explains why multi-objective training with human feedback converges to generic outputs like the ``GPT-3 wedding effect". Organizations decay in procedural complexity. Biological systems operate through hierarchy formation and extreme specialization at coordination boundaries~\cite{west1997general}. Recent empirical work has observed coordination difficulties in multi-agent systems~\cite{Rodriguez2016,Sowinski2022}, including oscillatory dynamics and consensus failures.

The topological extension~\cite{Chichilnisky1980} of Arrow's impossibility theorem~\cite{Arrow1963} applies when $L(P)$ exceeds collective working memory as the systems must always make a decision. Multi-objective optimization faces the same impossibility: no consistent aggregation of conflicting gradients exists, forcing cyclic behavior~\cite{Liu2021} analogous to voting paradoxes. Cycling could be the least worst option among chaotic wandering, convergence to limit cycles, or other dynamics. The required information needed to explore the phase space in the absence of an optimal aggregation method exceeds system capacity.

\subsection{Limitations}
TCT is descriptive of statistical trends, and not prescriptive about exact patterns. It implies that the dynamics are bounded, but cannot determine exact trajectories due to Level II chaos. 
It is possible that instead of avoiding coordination overheads, systems converge towards simplification due to selection pressure for robustness, generalization, or ease of transmission. However, other selection pressures remain qualitatively similar to coordination and most likely have their own information-theoretical bottlenecks.

All systems which successfully survive and propagate most likely discover
workable ways around paying the full $N^2d^2$ coordination costs, such as using hierarchical coordination protocols (Equation \ref{eq:hier}).

Further, effective $N$, $K$, $\varepsilon$, $d$, and utility change dynamically and finding them is not straightforward, which limits the calculation of other parameters which depend on them.

\subsection{Suggested System Design Strategies}
Systems face an irreducible trade-off between exploration (scaling with $N$ and $d$) and coordination costs (scaling as $N^2d^2$). Any meta-strategy to resolve this trade-off most likely adds an objective, worsening the original problem. Even the minimum set of objective inherent in the formulation of this framework has $d=2$: exploration and coordination, as both are needed to survive environmental fluctuations. This creates the paradox that low $N$, $d$ systems lack exploratory capacity while high $N, \ d$ systems face prohibitive coordination costs. Further, coordination attempts push the whole problem up one level of emergent complexity.

These results suggest that current approaches to AI alignment, organizational management, and social coordination may be fighting thermodynamics. Understanding these bounds enables design improvements within constraints rather than attempts to transcend them. Most natural systems maintain accuracy to some extent against coordination pressure through explicit resource allocation or modular architectures that minimize effective $N$ and $d$ at coordination boundaries, and have mitigation strategies to pay the hierarchical prediction errors which emerge.

Building on Arrow's impossibility theorem, any coordination mechanism must violate at least one fundamental condition. Systems can restrict universal domain by requiring single-peaked preferences, abandon Pareto efficiency by overriding unanimous preferences, violate independence of irrelevant alternatives (IIA) through ranked-choice voting where additional options affect binary comparisons, or implement single-agent determination through designated decision-makers. Each violation creates a distinct coordination focal point with compromises: domain restriction reduces expressiveness, abandoning Pareto efficiency permits collectively suboptimal outcomes, violating IIA enables path-dependent outcomes through ordering effects, and single-agent determination concentrates all influence in one decision node. Rather than allowing these violations to emerge through system failure, conscious selection enables designers to match the sacrificed principle to system requirements. Note that selecting which condition to violate itself faces Arrow's impossibility as there exists no consistent meta-procedure for choosing among violation strategies. 

Further, TCT shows that systems inevitably simplify to focal coordination points that are findable rather than optimal. However, given the network effects of Theorem 2, this could enable better design principles to work within these constraints. The suggested minimum design protocol to capitalize on the feedback loop created by reflexive dynamics is to ask ``\textit{What is the current focal point?}'' at each transition which necessitates simplification, as well as periodically to check for lock in effects decoupled from environmental feedback. Maintaining parallel hierarchies and redundant non-dominant focal points, even if they are comparable in model complexity, could enable rapid switching when the environment inevitably shifts.

\subsection{Conclusions}
We have derived fundamental bounds on multi-objective coordination accuracy in classical multi-agent systems from a combination of information theory, social choice theory, and statistical mechanics. The core result is that the information needed for coordination scales as $L(P) \geq NK\log K (1-\rho) + \binom{N}{2} \frac{d(d+3)}{2} \log(1/\varepsilon)$, but every algorithm to reach an optimal solution faces Arrow's Impossibility, and as the system collectively always decides something, the selection pressure for findability across agents far exceeds accuracy. This implies that large systems cannot maintain accuracy across multiple objectives. 

The framework unifies diverse phenomena through focal point selection. Whether in AI alignment, organizational decay, market dynamics, or scientific paradigm shifts, the pattern is identical: systems sacrifice accuracy for coordination when $L(P)$ exceeds available resources. Future work will develop this framework further and build upon domain specific applications.

For multi-agent multi-objective classical information-processing systems, TCT suggests a universal principle: more leads to the emergence of simplified focal points, which is different.

\ack{The author acknowledges relevant contributions across disciplines which could not be cited due to finite coordination capacity. The author thanks colleagues for helpful discussions, feedback, literature suggestions, and guidance. Beyond citations, some of these ideas have been influenced by the works of Terry Pratchett, Jared Diamond, and Philip Kapleau.}

\data{No new data was created or analyzed in this study.}

\suppdata{Supplementary PDF includes the topological derivation of the Theorem 1, derives the minimum description length of communication protocols in hierarchical coordination, and details characterization of Level II chaos dynamics using the bistromathics (restaurant bill splitting) example.}

\bibliography{gfep}
\bibliographystyle{unsrturl}

\end{document}


\articletype{Paper} 

\title{Supplementary for ``Coordination Requires Simplification: Thermodynamic Bounds on Multi-Objective Compromise in Natural and Artificial Intelligence''}

\author{Atma Anand$^1$\orcid{0000-0003-0616-7955}}

\affil{$^1$Department of Physics and Astronomy, \\ University of Rochester, Rochester, NY USA}

\email{atma.anand@rochester.edu}

\keywords{Scaling laws of complex systems, Thermodynamics of Computation, Information Theory, Statistical Mechanics, Chaos \& nonlinear dynamics, Artificial Intelligence, Social Physics}

\date{\today}

This supplementary material includes the topological derivation of the Theorem 1 in the main text, elaborates on the minimum description length of communication protocols in hierarchical coordination, and details characterization of Level II chaos dynamics using the bistromathics (restaurant bill splitting) example.

\section{Supplementary: Topological Derivation of Coordination Scaling}\label{sec:topo}
We sketch an extension of the topology of spaces of preferences \cite{Chichilnisky1980} and rederive the conflict resolution $\sim N^2d^2 \log(1/\varepsilon)$ bits needed in Theorem 1 through geometric reasoning.

\subsection{Preference Spaces and Their Topology}

Consider $d \geq 2$ objectives that agents must coordinate over. Each agent $i$ has continuous preferences represented by a utility function $u_i: \{1,2,...,d\} \rightarrow \mathbb{R}$.

The space of all possible preference orderings forms a \textit{preference space} $\mathcal{P}_d$. With continuous utilities, this becomes an infinite-dimensional space that inherits \textit{non-trivial topology} even for $d = 2$:


\begin{itemize}
    \item Multiple connected components corresponding to different preference orderings
    \item Non-trivial fundamental group (contains cycles that cannot be continuously deformed to a point)
    \item These cycles arise from the transitivity requirement: if A$>$B and B$>$C, then A$>$C must hold
\end{itemize}

The ``holes" in preference space represent fundamental conflicts where no continuous path exists between certain preference configurations while maintaining transitivity. These topological obstructions are what make continuous, fair aggregation impossible.

\textit{Non-Contractibility:} 
Specifically, $\mathcal{P}_d$ contains cycles that cannot be continuously shrunk to a point.

\textit{Example for $d=2$}: With two objectives, say preserving wealth and seeking high returns while investing. Consider evaluating three stock alternatives A, B, C. A preference cycle might be:
\begin{itemize}
    \item A: high on safety, low on payoff
    \item B: medium on both objectives  
    \item C: low on safety, high on payoff
\end{itemize}

\begin{figure*}[!htbp]
\centering
\includegraphics[width=\textwidth]{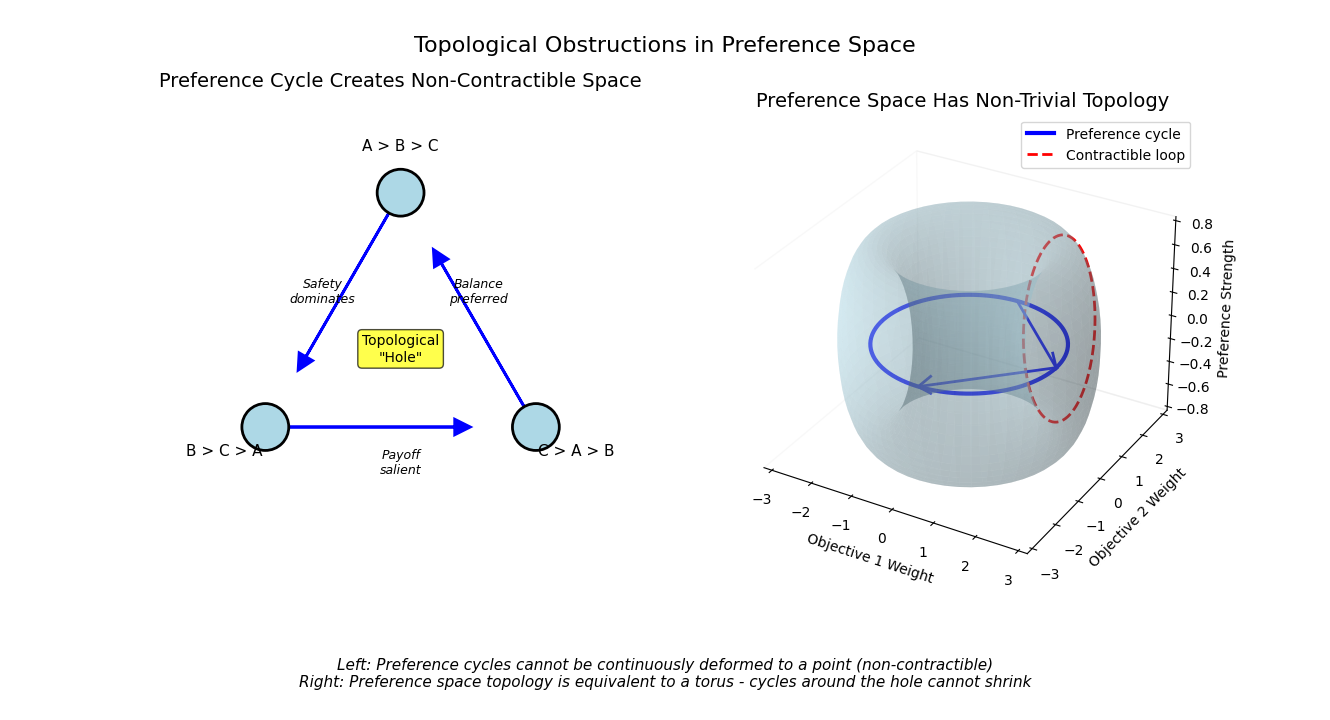}
\caption{\textbf{Topological Obstructions in Preference Space.} Left: Preference cycle $A>B>C>A$ emerges when agents evaluate options across multiple objectives. Different objectives dominate pairwise comparisons: safety considerations favor $A>B$, balanced criteria prefer $B>C$, and payoff maximization selects $C>A$. This creates a non-contractible loop around a topological ``hole'' in preference space. Right: The preference space has non-trivial topology equivalent to a torus. The blue curve shows the preference cycle that cannot be continuously deformed to a point, while the red dashed curve shows a contractible loop. This topological obstruction shows no consistent preference aggregation exists (Chichilnisky's extension of Arrow's theorem), forcing multi-objective systems to cycle rather than converge. This geometric constraint contributes to the $N^2d^2$ scaling in coordination complexity as each agent pair must reconcile incompatible preference topologies.}
\label{fig:topology_holes}
\end{figure*}

An agent weighting safety heavily prefers A$>$B$>$C. One weighting payoff prefers C$>$B$>$A. One weighting both equally prefers B$>$A and B$>$C. These create cycles in preference space that represent the fundamental tradeoffs between objectives or topological ``holes''. Figure \ref{fig:topology_holes} summarizes this.

\subsection{Chichilnisky's Impossibility Result}

Chichilnisky considered continuous aggregation functions that map N individual preferences to a collective preference. Even with continuous (infinite) preference options, certain fundamental conflicts cannot be resolved without forcing compromises.

\textbf{Theorem (Chichilnisky, 1980)}: For $d \geq 2$ objectives evaluating 3 or more alternatives, and $N \geq 2$ agents, there exists no continuous function $F: \mathcal{P}_d^N \rightarrow \mathcal{P}_d$ that satisfies:
\begin{enumerate}
\item \textit{Anonymity}: $F(u_{\sigma(1)}, ..., u_{\sigma(N)}) = F(u_1, ..., u_N)$ for any permutation $\sigma$
\item \textit{Unanimity}: If all $u_i$ agree on $x > y$, then $F(u_1, ..., u_N)$ has $x > y$
\item \textit{Continuity}: $F$ is continuous with respect to the natural topology
\end{enumerate}

Chichilnisky's formulation maps to Arrow's~\cite{Arrow1963} as follows: universal domain is implicit in the topological preference space $\mathcal{P}_d$, non-dictatorship is strengthened to anonymity, Pareto efficiency is equivalent to unanimity, and Independence of Irrelevant Alternatives (IIA) is replaced by continuity. This topological reformulation demonstrates that preference aggregation impossibility is robust to different axiomatic choices.


\subsubsection*{Application to Multi-Objective ML}

Chichilnisky's topological impossibility applies directly to gradient-based multi-objective optimization. Each objective loss function, $L_i(\theta)$ induces a preference ordering over parameter space $\theta \in \Theta$: for $\theta_1, \theta_2 \in \Theta$, we say $\theta_1 \succ_i \theta_2$ if $L_i(\theta_1) < L_i(\theta_2)$. The gradient $\nabla L_i(\theta)$ at each point defines a local preference direction. When objectives conflict with $\nabla L_i \cdot \nabla L_j < 0$, improvement in $L_i$ requires worsening $L_j$, creating cyclic preferences over the Pareto front: $\theta_A \succ_i \theta_B \succ_j \theta_C \succ_k \theta_A$. This constructs a non-contractible loop in the preference space analogous to the topological obstructions in Figure~\ref{fig:topology_holes}.

Standard approaches combine gradients as $\nabla\theta = \sum \alpha_i \nabla L_i$, where selecting weights $\{\alpha_i\}$ constitutes the aggregation function $F$. This gradient aggregation defines a social welfare function satisfying: (i) Unanimity: if $\nabla L_i$ all point in direction $v$, then $\nabla\theta \propto v$; (ii) Anonymity: invariant under permutation of loss indices; (iii) Continuity: $\alpha_i(\nabla L_1,...,\nabla L_d)$ varies continuously with gradients. Chichilnisky's theorem applies when the space of gradient configurations admits preference cycles (non-contractible topology), proving no such continuous, anonymous, unanimous aggregation exists. This forces either cycling behavior~\cite{Yu2020, Liu2021} or arbitrary weight selection that changes under reparametrization; not an algorithmic limitation but a topological necessity.

The impossibility operates at two levels in practice. First, at the gradient level analyzed above, conflicting objectives create non-convergent dynamics. Second, when multiple stakeholders (users, developers, safety teams) must collectively determine the loss weights themselves, they face an additional layer of impossibility: their preferences over the contractible weight simplex $\{\alpha_i \geq 0, \sum \alpha_i = 1\}$ inherit Chichilnisky-Arrow impossibility constraints directly. Recent work on aggregating human preferences for AI decision-making~\cite{noothigattu2018voting} attempts to navigate this space through voting mechanisms, which necessarily violate at least one of Chichilnisky's axioms (typically continuity or anonymity). The prevalence of dictatorial weight selection in deployed systems reflects this impossibility: absent a principled aggregation method, practitioners default to single-agent determination.

\subsection{From Topology to Information}

The impossibility arises from fundamental conflicts between objectives that force compromises.

To approximate the impossible continuous map to precision $\varepsilon$, we must specify:

\textit{Per compromise}: $\log_2(1/\varepsilon)$ bits to desired resolution

\textit{Per agent}: Total information to navigate all compromises across $d$ objectives needs pairwise comparison to precision $\varepsilon$:

$I_{\text{agent}} = \binom{d}{2}\log_2(1/\varepsilon) = \frac{d(d-1)}{2} \log_2(1/\varepsilon)$


Each compromise corresponds to a pair of objectives that cannot be simultaneously maximized, creating a topological obstruction in preference space as shown in Figure \ref{fig:topology_holes}. While continuous preferences create infinite variations of each conflict, we count only the fundamental basis, i.e., the d(d-1)/2 pairwise conflicts from which all higher-order conflicts arise. This represents the ``basis" or the minimum spanning set for the theoretical minimum information required for coordination.

\subsubsection{Multi-Agent Coordination Requirements}

With $N$ agents, agents must specify the raw weight and variance of their preferences pairwise (compare to mean and variance of the Gaussian model):
\begin{equation}
    I_{\text{w}} = \binom{N}{2} 2d \, \log_2(1/\varepsilon) = N(N-1)d \, \log_2(1/\varepsilon)
\end{equation}

\textit{Pairwise Coordination}
Each pair of agents must coordinate their navigation around each topological obstruction to maintain consistency (compare to off diagonal elements of the covariance matrix of the Gaussian model):
\begin{align}
I_{\text{conflict}} = \binom{N}{2} \frac{d(d-1)}{2} \log_2(1/\varepsilon) = \frac{N(N-1)d(d-1)}{4} \log_2(1/\varepsilon)
\end{align}

\textit{Model Complexity Term}: The coordination protocol must also specify how agents' internal models relate, contributing (same as the information-theoretical proof):

$$I_{\text{models}} = N \bar{K} \log \bar{K}  \,h(\rho)$$

where $\bar{K}$ is mean model complexity and $h(\rho)=1-\rho$ captures model diversity.

\subsubsection{Total Information Requirement}

Combining all terms:
\begin{align}
L(P) &\geq I_{\text{models}} + I_{\text{weight}} + I_{\text{conflict}} \nonumber \\
&= N \bar{K} \log \bar{K}  \,h(\rho) + N(N-1)d \log_2(1/\varepsilon) + \frac{N(N-1)d(d-1)}{4} \log_2(1/\varepsilon)\\
&= N \bar{K} \log \bar{K} \, h(\rho) + \frac{N(N-1)}{2} \frac{d(d+3)}{2} \log_2(1/\varepsilon) \nonumber
\end{align}
which is the same as Theorem 1 in the main paper.

\subsection{Interpretation}

The topological derivation reveals \textit{why} coordination conflict resolution requires $\mathcal{O}(N^2 d^2)$ information:
\begin{itemize}
\item Preference spaces contain $\frac{d(d-1)}{2}$ fundamental obstructions
\item Each obstruction requires $\log_2(1/\varepsilon)$ bits to resolve
\item $N$ agents create $\binom{N}{2}$ pairwise coordination requirements
\item Total scaling: $\binom{N}{2} \frac{d(d+3)}{2} \log_2(1/\varepsilon) \sim \mathcal{O}(N^2 d^2 \log(1/\varepsilon))$
\end{itemize}

This suggests that the information-theoretic bound in the main text reflects fundamental topological constraints, not algorithmic limitations. The convergence of these independent derivations strengthens the conclusion that coordination complexity scaling is quadratic.

\section{Hierarchical Coordination}\label{sec:hier}
We provide detailed analysis of the description length of communication protocols in hierarchical structures and show the tradeoffs involved in overcoming the $N^2d^2$ scaling limitation.

\subsection{Single-Level Hierarchy}
Consider $N$ agents organized into $M$ groups of size $N/M$. Following the structure of Theorem 1, in-group coordination requires:
\begin{equation}
L_{\rm{in}} = N \bar{K}_{\rm{in}} \log\bar{K}_{\rm{in}} h(\rho_{\rm{in}}) + 
M\binom{N/M}{2} \frac{d_{\rm{H}}(d_{\rm{H}}+3)}{2} \log(1/\varepsilon)
\end{equation}
Between or out-group coordination among $M$ representatives requires:
\begin{equation}
L_{\rm{out}} = M\bar{K}_{\rm{out}}\log\bar{K}_{\rm{out}} h(\rho_{\rm{out}}) + \frac{M(M-1)}{2} \frac{d_{\rm{H}}(d_{\rm{H}}+3)}{2} \log(1/\varepsilon)
\end{equation}
Total description length: $L_{\rm{H}} = L_{\rm{in}} + L_{\rm{out}}$

If we assume the in-terms equal the out-terms, this changes to:
\begin{align}
L_{\rm{H}} = (N + M)\bar{K}_{\rm{H}} \log\bar{K}_{\rm{H}}  h(\rho_{\rm{H}}) + \left[\frac{N^2 - NM}{2M} + \frac{M(M-1)}{2}\right] \frac{d_{\rm{H}}(d_{\rm{H}}+3)}{2} \log(1/\varepsilon)
\end{align}

Taking $\frac{\partial L_{\rm{H}}}{\partial M} = 0$ and assuming the second term dominates communication:
\begin{equation*}
\frac{\partial}{\partial M}\left[\frac{N^2 }{2M} - \frac{N}{2} + \frac{M(M-1)}{2}\right] = -\frac{N^2}{2M^2} + M - \frac{1}{2} = 0 
\end{equation*}

For reasonably large $N$ and $M$:
\begin{equation}
-\frac{N^2}{2M^2} + M \approx 0 \implies M_{\rm{opt}} = \frac{N^{2/3}}{2^{1/3}} \approx N^{2/3}
\end{equation}

Substituting back:
\begin{equation}
L_{\rm{H}}^{\rm{min}} \approx N\bar{K}_{\rm{H}}\log\bar{K}_{\rm{H}} h(\rho) + N^{4/3}\frac{d_{\rm{H}}(d_{\rm{H}}+3)}{2}\log(1/\varepsilon)
\end{equation}

\subsection{Multi-Level Hierarchy}
For a complete tree with branching factor $b$ and depth $D = \log_b(N)$, at level $\ell$ there are $n_\ell = N/b^\ell$ nodes, each node has internal coordination cost:
\begin{equation}
c_\ell = b\bar{K}\log\bar{K} h(\rho) + \frac{b(b-1)}{2} \frac{d_{\rm{H}}(d_{\rm{H}}+3)}{2}\log(1/\varepsilon)
\end{equation}

Summing across all levels:
\begin{equation}
L_{\rm{multi}} = \sum_{\ell=0}^{\log_b(N)-1} \frac{N}{b^\ell}  c_\ell 
\end{equation}

Since agents communicate once at the base of the hierarchy and then each hierarchy representatives shares their aggregate model of the hierarchy within the confines of Arrow's impossibility, in the limit of large $N$ the complexity term sums to:
\begin{equation}
\sum_{\ell=0}^{\log_b(N)-1} \frac{N}{b^\ell} b \bar{K}\log\bar{K} h(\rho) \approx N\bar{K}\log\bar{K} \, h(\rho)  \frac{b^2}{b-1}
\end{equation}
with the caveat that the $\bar{K}$ is extremely likely to saturate at the working memory capacity ($B$) at each boundary.

The communication term, using the geometric series $\sum_{\ell=0}^{D-1} b^{-\ell} = \frac{b(1-N^{-1})}{b-1}$:
\begin{align}
\sum_{\ell=0}^{\log_b(N)-1} \frac{N}{b^\ell}  \frac{b(b-1)d_{\rm{H}}(d_{\rm{H}}+3)}{4}\log(1/\varepsilon) = \frac{Nb^2d_{\rm{H}}(d_{\rm{H}}+3)}{4}\log(1/\varepsilon)  (1-N^{-1})
\end{align}

For large $N$:
\begin{align}
L_{\rm{multi}} \approx  N\bar{K}\log\bar{K}  \,h(\rho)  \frac{b^2}{b-1} + \frac{1}{4}Nb^2d_{\rm{H}}(d_{\rm{H}}+3)\log(1/\varepsilon)
\end{align}

This achieves $\mathcal{O}(N)$ for both the model and communication term, $\bar{K}$ typically saturates at working memory capacity $B$, and typically $d_{\rm{H}} < d$ . However, this analysis assumes: (1) Perfect information aggregation at each level with no loss, (2) Fixed model complexity $K(\approx B)$ and objective count $d_{\rm{H}}$ across hierarchical levels, (3) Zero inter-level coordination overhead. These assumptions effectively bypass Arrow's impossibility theorem by assuming perfect preference aggregation, which is violated in practice. Information theory requires that perfectly summarizing $b$ agents' states into one representative requires at least the mutual information between them, which scales with agent model variance. Thus, while hierarchies can drastically reduce coordination costs, it transforms the problem into information loss at boundaries, where simplification to focal points becomes thermodynamically necessary. The prevalence of hierarchical organization in biological systems~\cite{west1997general} suggests it may provide efficiency advantages that could include reduced coordination costs.

\section{Level II Chaos: How Coordination Changes the System}\label{sec:lvii}
The restaurant bill example exhibits Level II chaos as attempts to coordinate fundamentally alter the coordination landscape itself. We analyze this through the evolution of system states and emergent feedback loops.

\subsection{Bistromathics as a Dynamical System}
Consider the state evolution with the same 10 bits needed per objective per agent, and 5 bits for each pairwise conflict resolution:

\textit{Initial State} (t=0): N=6 diners, simple split assumed
\begin{itemize}
\item System state: $S_0 = \{N=6, d=1, T=0\}$ where everyone expects even split
\item Coordination cost: $L(P) = 6 \times 10 = 60 $ bits (Note: same 10 bit model for each diner, so $\rho \rightarrow 1$, but we ignore the resulting reduction)
\item Focal point: ``Split evenly"
\end{itemize}

\textit{Perturbation} (t=1): One diner mentions they only had salad
\begin{itemize}
\item Triggers cascade: $d: 1 \rightarrow 4$ (items, fairness, dietary restrictions, wealth)
\item New state: $S_1 = \{N=6, d=4, T=1\}$ 
\item Coordination cost (Theorem 1): $L(P) = 6 \times 40 + 15 \times 14 \times 5 = 1,290$ bits
\item System now unstable: $L(P) > $ human working memory capacity (4$\pm$1 chunks or $\mathcal{O}(100)$ bits depending on encoding \cite{Cowan2015})
\end{itemize}

\textbf{Level I Response} (t=2): Attempt itemized split
\begin{itemize}
\item 3 diners start calculating, 3 give up
\item Effective $N$ splits: $N_{active} = 3$, $N_{passive} = 3$
\item Creates two subsystems with different objectives
\item Temperature rises: $T_{co} \propto \text{variance}(m_i)$ increases
\end{itemize}

\textbf{Level II Chaos} (t=3): Coordination changes environment
\begin{itemize}
\item Waiter, observing chaos, announces: ``I need this table"
\item Time constraint added: $\tau_{max} = 5$ minutes
\item Someone suggests: ``Salad guy pays separately, the rest just split evenly"
\item New focal point emerges from coordination failure itself
\end{itemize}

\subsection{Mathematical Framework for Environmental Feedback}

Unlike systems with rigid Markov blankets, coordination exhibits fluid boundaries: $N$ changes, e.g. old diners skip in frustration or last minute plan changes, $d$ morphs as new constraints emerge, e.g. ``I'm gluten-free'', and cluster membership shifts with each failed attempt. The cycling frequency becomes non-stationary, making precise prediction impossible. This framework might be developed further in future work.

We define the system state: 
$$\mathbf{S}(t) = \{N(t), d(t), T_{co}(t), K, \varepsilon, \mathcal{E}(t)\}$$
where $\mathcal{E}(t)$ represents environmental conditions modified by coordination attempts. These particular forms are selected for the lower bounds while maintaining symmetry. We assume $K$ and $\varepsilon$ are independent of time for simplicity.

\textbf{Evolution Equations}:
\begin{align}
\frac{dN}{dt} &= -\gamma_N \cdot \mathbb{I}[L(P) > L_{critical}=MN] \\
\frac{dd}{dt} &= \alpha \cdot \text{conflicts}(t) - \beta \cdot \text{resolutions}(t) \\
\frac{dT}{dt} &= \frac{1}{N \bar{K}^2}\sum_{i,j} ||m_i - m_j||^2 - \lambda T \\
\frac{d\mathcal{E}}{dt} &= f(\text{coordination attempts}, \text{external pressure})
\end{align}

Where:
\begin{itemize}
\item $\gamma_N$: Rate of agent dropout when coordination fails
\item $\alpha, \beta$: Rates of objective emergence and resolution
\item $\lambda$: Cooling rate when coordination succeeds
\item $\mathbb{I}[x]$: Indicator function, is defined as:

\begin{equation}
    \mathbb{I}[x] = \begin{cases} 
    1 & \text{if } x \text{ is true} \\ \nonumber
    0 & \text{if } x \text{ is false}
\end{cases}
\end{equation}
\end{itemize}

\subsection{Focal Point Dynamics with N and d}

The selected focal point depends on the coordination pressure. Under the optimistic assumption of linear contributions and $K$ saturating at system working memory capacity, $K_{\rm{max}}=B$:

\textbf{Focal Point Complexity}:
$$K_{focal}(N,d) = B  \exp\left(-\frac{L(P)}{L_{capacity}}\right)$$

where:
$$L(P) \approx aN + bd^2 + cN^2 + eN^2d^2$$

For the bistromathics example with linear approximation (coefficients can be found by comparing to Theorem 1):
\begin{itemize}
\item Each additional diner: $+a$ bits 
\item Each additional objective: $+bd$ bits
\item Pairwise coordination: $+cN$ bits per person [Dominates for small $d$]
\item Full interaction: $+ed^2$ bits per person-pair
\end{itemize}

Note: Even with optimistic linear scaling, the $N^2$ term dominates for $N > 5$, explaining why groups larger than 6 inevitably converge to ``split evenly" regardless of initial intentions.

\noindent \textbf{Sample Focal Point Transitions}:
\begin{align*}
\text{For } & L(P) < 0.3 L_{capacity}: \ \text{Exact calculation} \\
\text{For } & 0.3 L_{capacity} < L(P) < 0.7 L_{capacity}: \ \text{Category-based} \\
\text{For } & L(P) > 0.7 L_{capacity}: \ \text{Equal split or separate checks}
\end{align*}

\noindent The linear model could be optimistic as it assumes:
\begin{itemize}
\item All agents contribute equally (ignores power dynamics)
\item All objectives combine additively (ignores nonlinear conflicts)
\item Perfect information sharing (ignores communication errors)
\end{itemize}
In reality:
\begin{itemize}
\item Transitions occur at lower N due to communication overhead
\item Power law effects: One ``difficult" person affects system like adding 2-3 average people
\item Threshold effects: System suddenly collapses rather than gradual degradation
\end{itemize}


\begin{table}[h]
\centering
\begin{tabular}{ccccc}
\hline
N \textbackslash \- d & d=1 & d=2 & d=3 & d=4 \\
\hline
N=2-3 & Exact & Exact & Category & Equal \\
  
N=4-5 & Exact & Category & Equal & Separate \\

N=6-8 & Simple & Equal & Separate & Chaos \\

N$>8$ & Equal & Separate & Chaos & Abandon \\
\hline
\end{tabular}
\caption{Sample focal point selection in bistromathics as function of (N,d). ``Exact" = precise calculation, ``Category" = split by meal type, ``Equal" = even split, ``Separate" = individual checks, ``Chaos" = no consensus and restaurant intervenes, ``Abandon" = someone pays all or table walks out}
\end{table}

\subsection{Feedback Loops and Emergent Hierarchy}

\textbf{Loop 1: Complexity-Dropout Feedback}
$$N \uparrow \Rightarrow L(P) \uparrow \Rightarrow \text{dropouts} \uparrow \Rightarrow N \downarrow$$
But dropouts change group dynamics, potentially increasing $d$.

\textbf{Loop 2: Failed Coordination $\rightarrow$ Environmental Change}
\begin{itemize}
\item Failed split $\rightarrow$ Impatient waiter $\rightarrow$ Time pressure
\item Time pressure $\rightarrow$ Forced focal point $\rightarrow$ Different optimization landscape
\item New landscape $\rightarrow$ Different coordination problem entirely
\end{itemize}

\textbf{Loop 3: Memory Effects}
\begin{align*}
    \text{Bad experience} \Rightarrow \text{Future behavior change} \Rightarrow \text{Modified pairwise model overlap }  (\rho)
\end{align*}
Agents with shared experiences develop correlated models, leading to spontaneous symmetry breaking into distinct clusters with high intra-cluster $\rho$ and low inter-cluster $\rho$. This emergent hierarchy shifts coordination to the inter-cluster level, e.g. ``Remember last time we tried to split precisely? Let's never do that again".

\subsection{Phase Space Analysis}

The system exhibits distinct regimes in $(N, d, T)$ space:

\textit{Regime I: Simple Coordination} $(N < 4, d = 1)$
\begin{itemize}
\item Exact calculation feasible
\item No chaos and perturbations dampen
\item Example: ``You had beer, I'll cover the wine"
\end{itemize}

\textbf{Regime II: Metastable} $(4 \leq N \leq 8, d = 2-3)$
\begin{itemize}
\item Coordination possible but fragile
\item Small perturbations can trigger cascade
\item Example: ``Wait, who ordered the caviar?"
\end{itemize}

\textbf{Regime III: Chaotic} $(N > 8$ or $d > 3)$
\begin{itemize}
\item Coordination attempts modify environment
\item Multiple feedback loops active
\item Forced focal points: ``Split evenly" or ``Separate checks"
\end{itemize}

\subsection{Signatures of Level II Chaos}

\begin{enumerate}
\item \textbf{Hysteresis}: Once simplified (e.g., ``split evenly"), groups resist returning to precise calculation even when $N$ decreases

\item \textbf{Punctuated Equilibria}: Long periods of stable focal points interrupted by rapid transitions when $L(P)$ exceeds critical threshold

\item \textbf{Environmental Coupling}: External factors (impatient waiter, closing time, social pressure) become dominant when internal coordination fails

\item \textbf{Emergent Constraints}: New rules emerge from failed coordination:
   \begin{itemize}
   \item ``Maximum 3 cards"
   \item ``20\% gratuity added for parties of 6+"
   \item ``Separate checks must be requested at ordering"
   \end{itemize}
\end{enumerate}

In Bistromathics, potential emergent adaptations include one diner covering the difference to avoid complexity ($N\rightarrow1$), or in perhaps the most unfalsifiable claim in this publication, one postdoc convincing a group of $\approx15$ potentially inebriated conference attendees into calculating exactly what they owed while a very patient server waited while the totals matched and billed everyone individually.

\subsection{Generalization Beyond Bistromathics}

This Level II chaos appears universally in complex adaptive systems coupled to the environment when coordination requirements $L(P)$ approach system capacity. Near that threshold, systems enter a regime where:
\begin{itemize}
    \item Coordination attempts change the environment, e.g. failed bill splitting $\rightarrow$ impatient waiter
    \item Environmental changes force new coordination, e.g. pressure from impatient waiter $\rightarrow$ ``just split evenly"
    \item These become locked in feedback loops with bidirectional causation
\end{itemize}

When $L(P)$ is sufficiently below system capacity, environment is fixed and coordination happens within it.
Above this threshold, coordination attempts and environment co-evolve chaotically.

Examples:
\begin{itemize}
\item \textbf{Neural networks}: Failed gradient coordination $\rightarrow$ Architecture changes
\item \textbf{Organizations}: Failed meetings $\rightarrow$ New meeting structures $\rightarrow$ Different coordination problems
\item \textbf{Markets}: Failed price discovery $\rightarrow$ Trading halts $\rightarrow$ New market rules
\item \textbf{Democracies}: Failed multi-party coordination $\rightarrow$ Electoral system changes
\end{itemize}

Systems don't just optimize within fixed environments, but optimization attempts reshape the environment itself, creating truly chaotic dynamics where prediction requires modeling both system and environmental evolution.

\bibliography{gfep}
\bibliographystyle{unsrturl}